\documentclass[10pt,twocolumn,letterpaper]{article}

\usepackage[T1]{fontenc}
\usepackage[english]{babel}

\usepackage{cvpr}
\usepackage{times}
\usepackage{epsfig}
\usepackage{graphicx}
\usepackage{amsmath}
\usepackage{amssymb}

\usepackage{booktabs}       
\usepackage{multirow}
\usepackage{xcolor}
\usepackage[export]{adjustbox}
\usepackage{amsthm}
\usepackage{url}            
\usepackage{gensymb}

\usepackage[nomessages]{fp}

\newtheorem{observation}{Observation}

\usepackage{color}

\newcommand*{\affmark}[1][*]{\textsuperscript{#1}}

\usepackage{rotating}

\usepackage[pagebackref=true,breaklinks=true,letterpaper=true,colorlinks,bookmarks=false]{hyperref}

\cvprfinalcopy 

\def\cvprPaperID{673} 

\begin{document}

\title{Quad-networks: unsupervised learning to rank for interest point detection}

\author{Nikolay Savinov\affmark[1], Akihito Seki\affmark[2], \v Lubor Ladick\' y\affmark[1], Torsten Sattler\affmark[1] and Marc Pollefeys\affmark[1,3]\\
\affmark[1]Department of Computer Science at ETH Zurich, \affmark[2]Toshiba Corporation, \affmark[3]Microsoft \\
{\tt\small \{nikolay.savinov,ladickyl,sattlert,marc.pollefeys\}@inf.ethz.ch, akihito.seki@toshiba.co.jp}
}

\maketitle

\begin{abstract}
Several machine learning tasks require to represent the data using only a sparse set of interest points. An ideal detector is able to find the corresponding interest points even if the data undergo a transformation typical for a given domain. Since the task is of high practical interest in computer vision, many hand-crafted solutions were proposed. In this paper, we ask a fundamental question: can we learn such detectors from scratch? Since it is often unclear what points are "interesting", human labelling cannot be used to find a truly unbiased solution. Therefore, the task requires an unsupervised formulation. We are the first to propose such a formulation: training a neural network to rank points in a transformation-invariant manner. Interest points are then extracted from the top/bottom quantiles of this ranking. We validate our approach on two tasks: standard RGB image interest point detection and challenging cross-modal interest point detection between RGB and depth images. We quantitatively show that our unsupervised method performs better or on-par with baselines.
\end{abstract}

\section{Introduction}

Machine learning tasks are typically subdivided into two groups: supervised (when labels for data are provided by human annotators) and unsupervised (no data labelled). Recently, more labelled data with millions of examples have become available (for example, Imagenet~\cite{russakovsky2015imagenet}, Microsoft COCO~\cite{lin2014microsoft}), which led to significant progress in supervised learning research. This progress is partly due to the emergence of convenient labelling systems like Amazon Mechanical Turk. Still, the human labelling process is expensive and does not scale well. Moreover, it often requires a substantial effort to explain human annotators how to label data.

Learning an interest point detector is a task where labelling ambiguity goes to extremes. In images, for example, we are interested in a sparse set of image locations which can be detected repeatably even if the image undergoes a significant viewpoint or illumination change. These points can further be matched for correspondences in related images and used for estimating the sparse 3D structure of the scene or camera positions. Although we have some intuition about what properties interest points should possess, it is unclear how to design an optimal detector that satisfies them. As a result, if we give this task to a human assessor, he would probably select whatever catches his eye (maybe corners or blobs), but that might not be repeatable.

In some cases, humans have no intuition what points could be "interesting". Let's assume one wants to match new images to untextured parts of an existing 3D model~\cite{Plotz_2015_ICCV}. The first step could be an interest point detection in two different modalities: RGB and depth map, representing the 3D model. The goal would be to have the same points detected in both. It is particularly challenging to design such a detector since depth maps look very different from natural images. That means simple heuristics will fail: the strongest corners/blobs in RGB might come from texture which is missing in depth maps.

Aiming at being independent of human assessment, we propose a novel approach to interest point detection via unsupervised learning. Up to our knowledge, unsupervised learning for this task has not yet been explored in previous work. Some earlier works hand-crafted detectors like DoG~\cite{lowe2004distinctive}. More recent works used supervised learning to select a "good" subset of detections from a hand-crafted detector. For example, LIFT~\cite{lift2016} aims to extract a subset of DoG detections that are matched correctly in the later stages of the sparse 3D reconstruction. However, relying on existing detectors is not an option in complicated cases like a cross-modal one. Our method, by contrast, learns the solution from scratch.

\begin{figure*}
 \centering
\includegraphics[width=\linewidth]{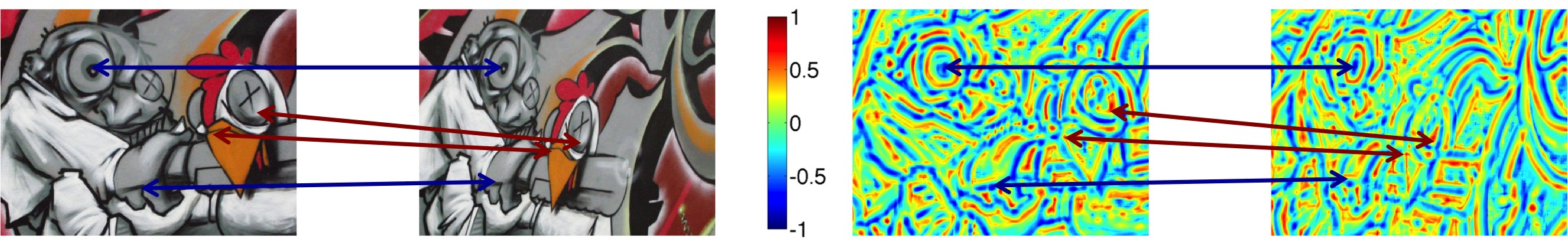}
 \caption{\label{fig:teaser} Left: an image undergoes a perspective change transformation. Right: our learned response function, visualized as a heat map, produces a ranking of image locations that is reasonably invariant under the transformation. Since the resulting ranking is largely repeatable, the top/bottom quantiles of the response function are also repeatable (examples of interest points are shown by arrows).}
 \end{figure*}

The idea of our method is to train a neural network that maps an object point to a single real-valued response and then rank points according to this response. This ranking is optimized to be repeatable under the desired transformation classes: if one point is higher in the ranking than another one, it should still be higher after a transformation. Consequently, the top/bottom quantiles of the response are repeatable and can be used as interest points. This idea is illustrated in Fig.~\ref{fig:teaser}.

When detecting interest points, it is often required to output not only the position of the point in the image, but also some additional parameters like scale or rotation. The detected values of those parameters are influenced by the transformations applied to the images. All transformations can be split into two groups based on their desired impact on the output of the detector. Transformations for which the detector is supposed to give the same result are called invariant. Transformations that should transform the result of a detector together with the transformation --- and thus their parameters have to be estimated as latent variables --- are called covariant~\cite{mikolajczyk2005comparison}. When learning a detector with our method, we can choose covariant and invariant transformations as it suits our goals. This choice is implemented as a choice of training data and does not influence the formulation.

The paper is organized as follows. In section~\ref{sec:related}, we discuss the related work. In section~\ref{sec:detByRank}, we introduce our formulation of the detection problem as the unsupervised learning to rank problem, and show how to optimize it. In section~\ref{sec:imagePoint}, we demonstrate how to apply our method to interest point detection in images. Finally, in section~\ref{sec:experiments} we validate our approach experimentally and conclude in section~\ref{sec:conclusion} by summarizing the paper and listing possibilities for future work.

\section{Related work}
\label{sec:related}

Currently, unsupervised learning comprises many directions: learning the distribution that best explains data (Gaussian Mixture Models learned via an EM-algorithm~\cite{mclachlan2004finite}, Restricted Boltzmann Machines~\cite{hinton2002training}, Generative Adversarial Nets~\cite{goodfellow2014generative}), clustering, dimensionality reduction and unsupervised segmentation (kMeans~\cite{hartigan1979algorithm}, LLE~\cite{LLE}, Isomap~\cite{Isomap}, PCA~\cite{jolliffe2002principal}, Normalized Cuts~\cite{shi2000normalized}, t-SNE~\cite{tSNE}), learning to simulate task solvers (when a solution is provided by the solver and the task is automatically generated~\cite{jeong2015data},~\cite{lerer2016learning}), and learning data representation suitable for further use in some other task (autoencoders~\cite{hinton2006reducing}, deep convolutional adversarial nets~\cite{radford2015unsupervised}, learning by context prediction~\cite{doersch2015unsupervised}, learning from tracking in videos~\cite{FeaturesFromVideos}, metric learning~\cite{wohlhart2015learning}, learning by predicting inpainting~\cite{ContextEncoders}, learning by solving jigsaw puzzles~\cite{JigsawPuzzles}).

While some tasks actually have a non-human label (for example, in solver simulation we can obtain the solution by running a solver), others (for example, representation learning) have none at all. Instead, they try to find an auxiliary task which is hard enough in order to learn a representation that is useful for already existing tasks (classification, for example). Designing such a task is non-trivial, therefore only few successful approaches exist (for example, \cite{doersch2015unsupervised}).

Our approach, on the other hand, does not require designing an unrelated auxiliary task. If we can obtain a repeatable ranking, then the top/bottom quantiles of this ranking can be used as detections.

One particular application of our method is interest point detection in images. Most of the existing image interest point detectors are hand-crafted to select particular visual elements like blobs, corners or edges. These include the DoG detector~\cite{lowe2004distinctive}, the Harris corner detector~\cite{harris1988combined} and its affine-covariant version~\cite{mikolajczyk2004scale}, the FAST corner detector~\cite{trajkovic1998fast} and the MSER detector~\cite{matas2004robust}. Most recently, there also emerged methods that do supervised learning building upon a hand-crafted solution: LIFT~\cite{lift2016} aims to extract an SfM-surviving subset of DoG detections, TILDE~\cite{verdie2015tilde} uses DoG for collecting the training set, \cite{lenc2016learning} samples training points only where LoG filter gives large absolute-value response. Building upon a hand-crafted detector restricts those supervised approaches to a subset of their basic method detections --- which makes those approaches inapplicable in the cases where there is no good detector yet. Our unsupervised method instead learns the detector completely from scratch by optimizing for a repeatable ranking.

Finally, a particularly challenging case in image interest point detection is the cross-modal one: the interest points should be repeatable among different image modalities. Several works mention this complex issue (\cite{Plotz_2015_ICCV}, \cite{aguilera2012multispectral}, \cite{kelman2007keypoint}, \cite{mishkin2015wxbs}) but do not propose a general solution. Our approach, on the contrary, is general in a sense that the same learning procedure could be applied to different tasks: we show it to work for RGB/RGB and RGB/depth modality pairs.

\section{Detection by ranking}
\label{sec:detByRank}

In this section we introduce the problem of learning an interest point detector as the problem of learning to rank points. We consider interest points to come from the top/bottom quantiles of some response function. If these quantiles are preserved under certain transformation classes, we have a good detector: it re-detects the same points. For the quantiles of the ranking to be preserved, we search for a ranking which is invariant to those transformations.

Let us consider a set $D$ of objects, every object $d \in D$ being an $N_d$-dimensional tuple of points $(p_d^1, \dots, p_d^{N_d})$. Each point $p_d^i$ comes from a set $P$ of points. Each object $d$ can undergo transformations from a set $T: D \mapsto D$. Each transformation $t \in T$ preserves certain point correspondences: some points in object $t(d)$ will correspond to points in object $d$. We assume one point can have at most one correspondence in the other object. To simplify the notation, we assume the correspondences have the same indexes in an object $d$ before and after a transformation. We denote the set of corresponding point indexes as $C_{dt} = \{ i_1, \dots, i_{K_{dt}} \}$, where $K_{dt}$ is the number of correspondences for points in $d$ and $t(d)$.

We want to rank object points and represent this ranking with a single real-valued response function $H(p | w)$, where $p \in P$ is a point and $w \in \mathbb{R}^n$ is a vector of parameters (one possible choice of $H$ is a neural network). Thus, invariance of the ranking under transformation $t \in T$ can be stated as follows: for every quadruple $(p_d^i, p_d^j, p_{t(d)}^i, p_{t(d)}^j)$ satisfying $i, j \in C_{dt}, i \neq j$, it holds that
\begin{align}
\begin{cases}
	H(p_d^i | w) > H(p_d^j | w)&\& \enspace\; H(p_{t(d)}^i | w) > H(p_{t(d)}^j | w) \\
	&\text{or} \\
	H(p_d^i | w) < H(p_d^j | w)&\& \enspace\; H(p_{t(d)}^i | w) < H(p_{t(d)}^j | w) \enspace.
\end{cases}
\label{eq:ineq}
\end{align}
From the condition above it follows that
\begin{observation}
	If $H$ satisfies the ranking constraints (\ref{eq:ineq}) and every point has a correspondence, the top/bottom quantiles of $H$ before a transformation correspond to the top/bottom quantiles of $H$ after it.
\label{ob:quantile}
\end{observation}

Thus, to get a repeatable interest point detector, one just needs to sort all points $p$ of the object $d$ by their response $H(p | w)$ and take the top/bottom quantiles as interest points.

In the next section, we will state the optimization objective aiming at preserving the ranking (\ref{eq:ineq}).

\subsection{Ranking objective and optimization}

First, let us introduce a ranking agreement function for quadruples:
\begin{align}
	R&(p_d^i, p_d^j, p_{t(d)}^i, p_{t(d)}^j | w) = \nonumber \\
	& (H(p_d^i | w) - H(p_d^j  | w)) (H(p_{t(d)}^i  | w) - H(p_{t(d)}^j  | w)) \enspace.
\end{align}
Then the ranking invariance condition~(\ref{eq:ineq}) can be re-written as
\begin{equation}
	R(p_d^i, p_d^j, p_{t(d)}^i, p_{t(d)}^j  | w) > 0 \enspace.
\label{eq:ranking}
\end{equation}
In order to give preference to this invariance, we will assume the object set $D$ and the transformation set $T$ to be finite (for the sake of training) and minimize the objective:
\begin{align}
	L(w) = \sum\limits_{d \in D} \sum\limits_{t \in T} \!\!\! \sum\limits_{\phantom{-}i, j \in C_{dt}} \!\!\!\!\! \ell(R(p_d^i, p_d^j, p_{t(d)}^i, p_{t(d)}^j  | w)) \enspace,
\end{align}
where $\ell(R)$ is a function penalizing non-positive values. One naive solution would be to use a "misranking count" loss
\begin{align}
\ell(R) =
\begin{cases}
	1, & \text{if}\ R \leq 0 \enspace, \\
	0, & \text{otherwise} \enspace.
\end{cases}
\end{align}
Unfortunately, this loss is hard to optimize as it either does not have a gradient or its gradient is zero. Instead, we upper-bound the discontinuous loss with a differentiable one. In this work, we choose to use the hinge loss
\begin{align}
\ell(R) = \max(0, 1 - R) \enspace.
\end{align}
Then the final form of our minimized objective will be
\begin{align}
	L(w) \! = \! \sum\limits_{d \in D} \sum\limits_{t \in T} \!\!\! \sum\limits_{\phantom{-}i, j \in C_{dt}} \!\!\!\!\! \max(0, 1 \! - \! R(p_d^i, p_d^j, p_{t(d)}^i, p_{t(d)}^j  | w)),
\label{eq:objective}
\end{align}
which is differentiable as long as $H(p  | w)$ is differentiable w.r.t $w$ (that is satisfied if $H$ is a neural network; note that the objective above is non-convex even if $H$ is linear). Therefore, we can use gradient descent for the optimization.

\section{Image interest point detector learning}
\label{sec:imagePoint}
Learning detectors from scratch is a hard task since it is non-trivial to formulate good detector criteria in the optimization framework. As investigated by~\cite{mikolajczyk2005comparison}, a good detector should produce interest points that are robust to viewpoint/illumination changes (to detect the same points and further match them) and sparse (to make feature matching feasible). To comply with the earlier introduced terminology, $d$ is an image, $p$ is a position in the image represented by a patch, a transformation $t$ is a viewpoint/illumination change and correspondence sets $C_{dt}$ are patch-to-patch correspondences between images observing the same 3D scene.

It is typical for interest point detectors to ensure sparsity in two ways: by retaining the top/bottom quantiles of the response function (contrast filtering) and by retaining the local extrema of the response function (non-maximum suppression). While Observation \ref{ob:quantile} suggests reproducibility of our detections under contrast filtering, it turns out that non-maximum suppression is also suitable for our detector according to the following
\begin{observation}
	If $H$ satisfies the ranking constraints (\ref{eq:ineq}) and the vicinity of the correspondence $(p_d, p_{t(d)})$ is visible in both images $d$ and $t(d)$, then $p_d$ is a local extremum in image $d$ $\iff$ $p_{t(d)}$ is a local extremum in image $t(d)$.
\label{ob:extrema}
\end{observation}
It is easy to see why this observation is true: if the position is ranked higher/lower than all the neighbors in one image, the corresponding position should be ranked higher/lower than the corresponding neighbors in another image.

Thus the proposed objective is beneficial for the detector pipeline which consists of non-maximum suppression and contrast filtering. This pipeline is followed by many detectors including the popular DoG detector~\cite{lowe2004distinctive}. In the following section, we explain how to train an image interest point detector with our objective. 

\subsection{Training}
\label{sec:training}
We need to sample from both the image set $D$ and the transformation set $T$ to perform training with the objective (\ref{eq:objective}). We could, of course, take images and transformations from any available image dataset with correspondences. But this does not address two important questions:
\begin{itemize}
	\item How to achieve invariance exactly to the transformations that we want? For example, most real images are taken up-right, so there is no relative rotation between any pair of them. But we want our detector to be robust to cases where there is such a rotation.
	\item How to augment the training images? For example, all the objects in the training images might be well-illuminated. But in the testing images some objects might be in the shadow, while others are in the light. We might want to be robust to such cases. 
\end{itemize}
In this chapter we will show how to achieve each goal by randomly transforming training quadruples
\begin{align}
Q = (p_d^k, p_d^m, p_{t(d)}^k, p_{t(d)}^m) \enspace.
\end{align}

To achieve invariance to a transformation class $T_i$, we can sample two random transformations $t_{i_1} \in T_i, t_{i_2} \in T_i$ and apply a quadruple of transformations $(t_{i_1}, t_{i_1}, t_{i_2}, t_{i_2})$ to the training quadruples $Q$ element-wise. This expresses our preference to preserve the ranking even if a random transformation from $T_i$ is applied to the image.

To augment the data with a transformation class $T_a$, we can sample two random transformations $t_{a_1} \in T_a, t_{a_2} \in T_a$ and apply $(t_{a_1}, t_{a_2}, t_{a_1}, t_{a_2})$ to $Q$. This means that we apply the same transformation to both patches in the correspondence to create more training data.

Finally, there are some invariant/augmenting transformations which can't be easily parametrized and sampled (\eg, the non-Lambertian effect). In that case, we fully rely on their distribution, coming from the real data.

\def \wf {0.7} 
\begin{figure*}
 \centering
 \includegraphics[width=\wf\linewidth]{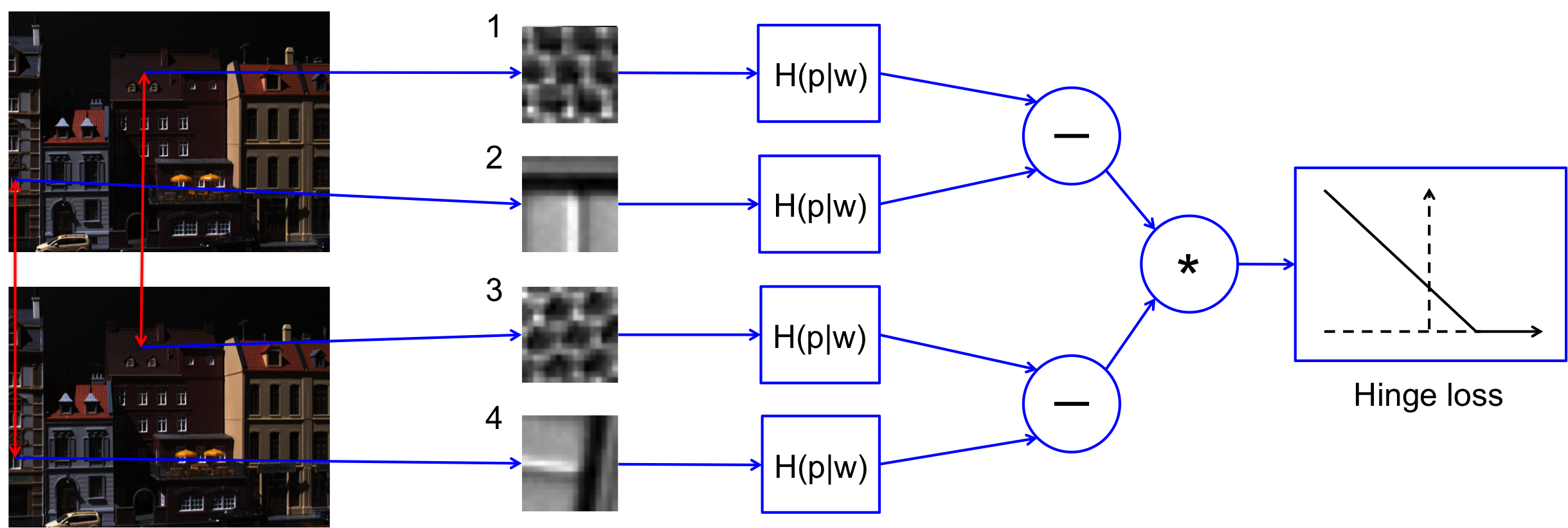}
 \caption{\label{fig:pipeline} Quad-network forward pass on a training quadruple. Patches $(1, 3)$ and  $(2, 4)$ are correspondence pairs between two different images, so $1,  2$ come from the first image and $3, 4$ come from the second image. All of the patches are extracted with a random rotation.}
 \end{figure*}

\section{Experiments}
\label{sec:experiments}
Our objective function (\ref{eq:objective}) is based on pairs of correspondences, forming training quadruples (an example of such a quadruple is shown in Fig.~\ref{fig:pipeline}). To train a detector, we need to obtain those correspondences. We investigated learning
\begin{itemize}
	\item an RGB detector from ground-truth correspondences (they come from projecting laser-scanned 3D points onto images),
	\item a fully-unsupervised RGB detector (correspondences are obtained by randomly warping images and changing illumination),
	\item a cross-modal RGB/depth detector (correspondences are trivially obtained as coinciding locations in view-aligned Kinect RGBD frames).
\end{itemize}
We further describe the setup of those experiments.

\noindent\textbf{Detector class.} We concentrate on the most commonly used type of detectors: scale-space-covariant, rotation-invariant ones (although our method is suitable for any combination of detector covariances/invariances). For example, DoG belongs to that type. Those detectors consider an interest point $p$ to be characterized by an image location $x, y$ and a scale $s$. The points are detected in a $3$-dimensional space (scale-space) using a response function
\begin{equation}
	H(p | w) = H(x, y, s | w) \enspace.
\end{equation}
Consequently, non-maximum suppression and contrast filtering work in this $3$-dimensional space as well (with a $3~\!\!\times\!\!~3~\!\!\times\!\!~3$ neighborhood).
Since rotation is not estimated, the detector is required to be invariant to it. The invariance is achieved by the random sampling (see Section~\ref{sec:training}).

\noindent\textbf{Detector evaluation.} DoG is the most widely used detector nowadays, so we use it as a baseline in our evaluation. The whole detector is a multi-stage pipeline in which we aim to substitute a crucial part: the filter used to convolve the image. In order to make a fair evaluation, we fix all the other stages of the pipeline. The whole procedure works as follows. First, we apply the response function $H(p)$ to all spatial positions of the image at all considered scales. This is the stage we are aiming to substitute with a learned function (DoG filter in the standard pipeline). Second, we do non-maximum suppression in scale-space. Third, we do accurate localization based on the second-order Taylor expansion of the response function around potential interest points~\cite{lowe2004distinctive}. Finally, we only take points for which the absolute value of the response is larger than a threshold.

For quantitative evaluation, we use the repeatability measure described in~\cite{mikolajczyk2005comparison} (with the overlap threshold parameter equal $40$\%). The repeatability is the ratio between the number of points correctly detected in a pair of images and the number of detected points in the image with the lowest number of detections. It is only meaningful to compare methods producing the same number of interest points: otherwise some method might report too many points and unfairly outperform others (\eg, if we take all points as "interesting", repeatability will be very high). Therefore, we consider a range of top/bottom quantiles, producing the desired numbers of points and compare all methods for those fixed numbers.

\noindent\textbf{Response function.}
In all experiments, the response function $H(p  | w)$ is a neural network. We describe it as a tuple of layers and use the notation:
\begin{itemize}
	\item $c(f, i, o, p)$ for convolutional layers with filter size $f \times f$, taking $i$ input channels, outputting $o$ channels, using zero-padding of $p$ pixels on each border (stride is always $1$ in all experiments),
	\item $f(i, o)$ for fully-connected layers, taking $i$ features and outputting $o$ features,
	\item $e$ for the ELU non-linearity function~\cite{clevert2015fast},
	\item $b$ for a batch normalization layer,
	\item $(\cdot)^n$ for applying the same network $n$ times.
\end{itemize}
In all the experiments, the response function is applied to grayscale $17$x$17$ patches. If the training data is in color, we convert it to grayscale. The patches are preprocessed as it is typical for neural networks: the mean over the whole patch is subtracted, then it is divided by the standard deviation over the patch.

\noindent\textbf{Augmentation.}
We augment the training data (see Section~\ref{sec:training}) with random rotations from $[0, 2 \pi]$ and random scale changes from $[\frac{1}{3}, 3]$.

\noindent\textbf{Optimization details.}
To optimize the objective (\ref{eq:objective}), we use the Adadelta algorithm~\cite{zeiler2012adadelta}, which is a version of gradient descent that chooses the gradient step size per-parameter automatically. We implement the model and optimization on a GPU (Nvidia Titan X) using the Torch7 framework~\cite{torch}. The batch size is $256$, our models are trained for $2000$ epochs, each consisting of randomly sampling a pair of corresponding images and then randomly sampling $10000$ quadruples from this pair. Eventually, by the time training stops our models have seen $20$ million sampled quadruples.

\subsection{RGB detector from ground-truth correspondences}
In this experiment, we show how to use existing 3D data to establish correspondences for training a detector. 

\noindent\textbf{Training.} We used the DTU Robot Image Dataset~\cite{dtu_robot}. It has 3D points, coming from a laser scanner, and camera poses, which allow to project 3D points into the pairs of images and extract image patches centered at the projections. Those projections form the correspondence pairs for training.

\noindent\textbf{Testing.} We used the Oxford VGG dataset~\cite{mikolajczyk2005comparison}, commonly chosen for this kind of evaluation. This dataset consists of $40$ image pairs.

\noindent\textbf{NN architectures.}
In this experiment, we tested two NN architectures: a linear model $(c(17, 1, 1, 0))$ and a non-linear NN with one hidden layer $(c(17, 1, 32, 0), e, f(32, 1))$.

\noindent\textbf{Results.}
We demonstrate that the filter of our learned linear model is different from the filters of the baselines in Fig.~\ref{fig:filters}. Furthermore, we show the detections of the linear model in comparison to DoG in Fig.~\ref{fig:detections}. Our learned model detects points different from DoG: they are more evenly distributed in images. That is usually profitable for estimating geometric transformations between camera frames.

The learned response functions with both investigated architectures (linear, non-linear) demonstrate better performance than baselines in most cases, as shown in Table~\ref{tab:3d} (results are averaged over all image pairs for each transformation type). Moreover, the non-linear model performs better than the linear one in the majority of the cases.

\begin{figure}[!ht]
 \centering
 \def \wlf {0.46}
\begin{tabular}{cc}
 \includegraphics[width=\wlf\linewidth]{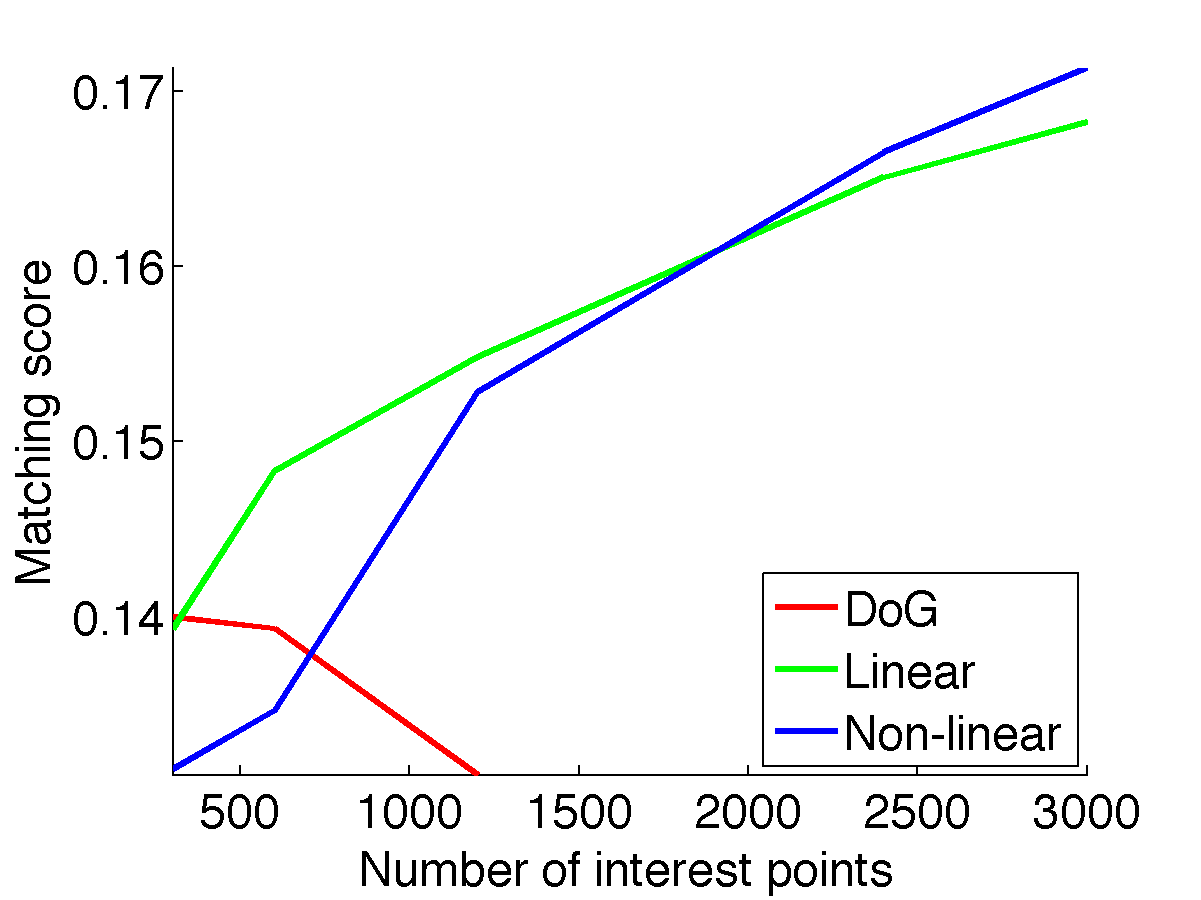} &
 \includegraphics[width=\wlf\linewidth]{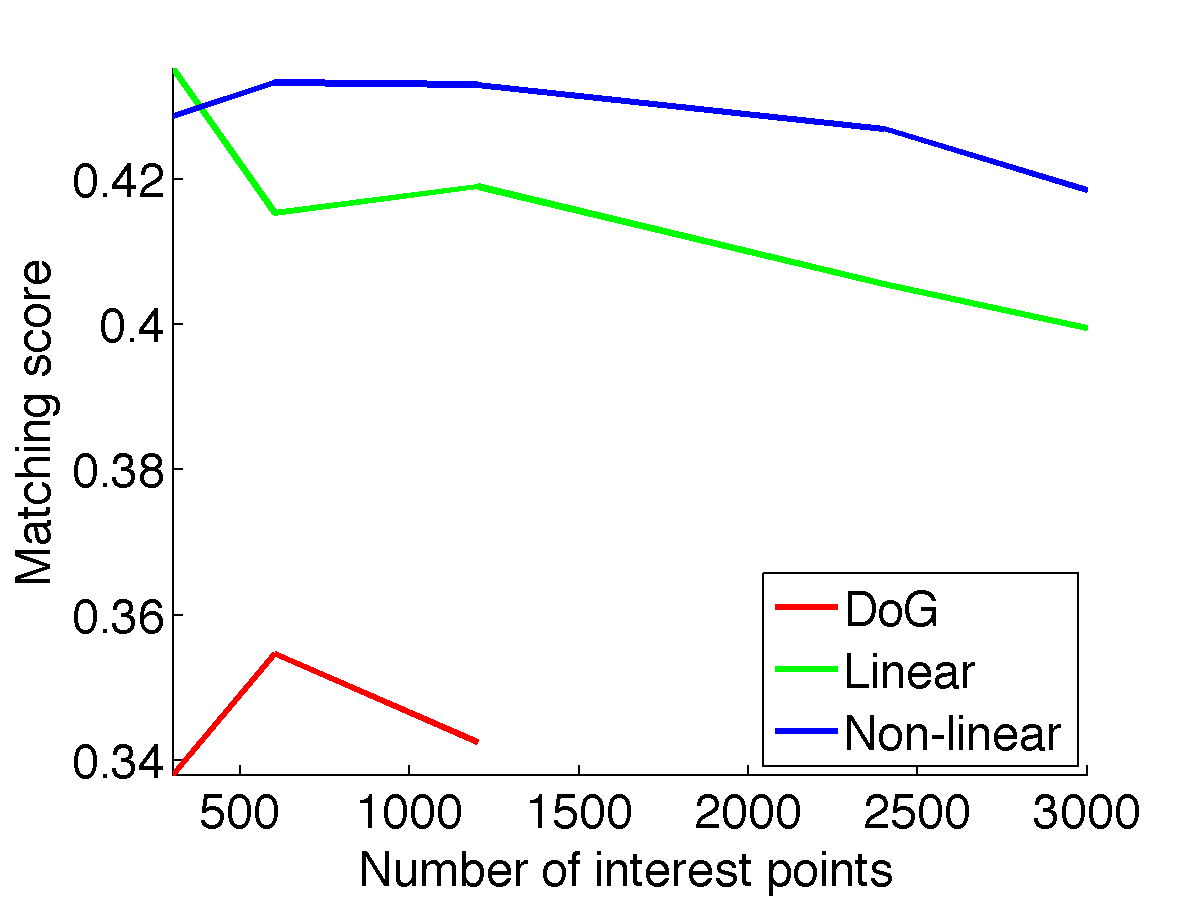} \\
 Wall (viewpoint) & Leuven (illumination) \\
 \includegraphics[width=\wlf\linewidth]{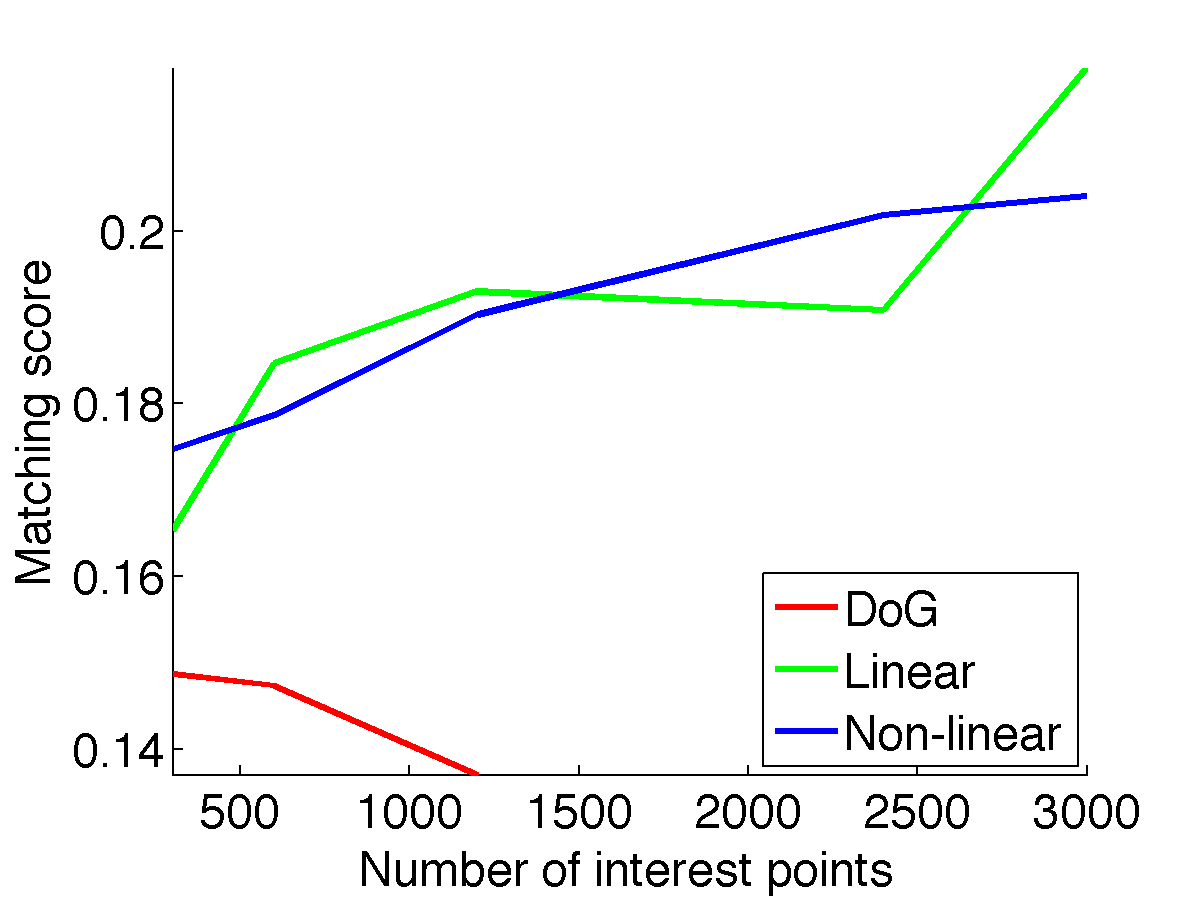} &
 \includegraphics[width=\wlf\linewidth]{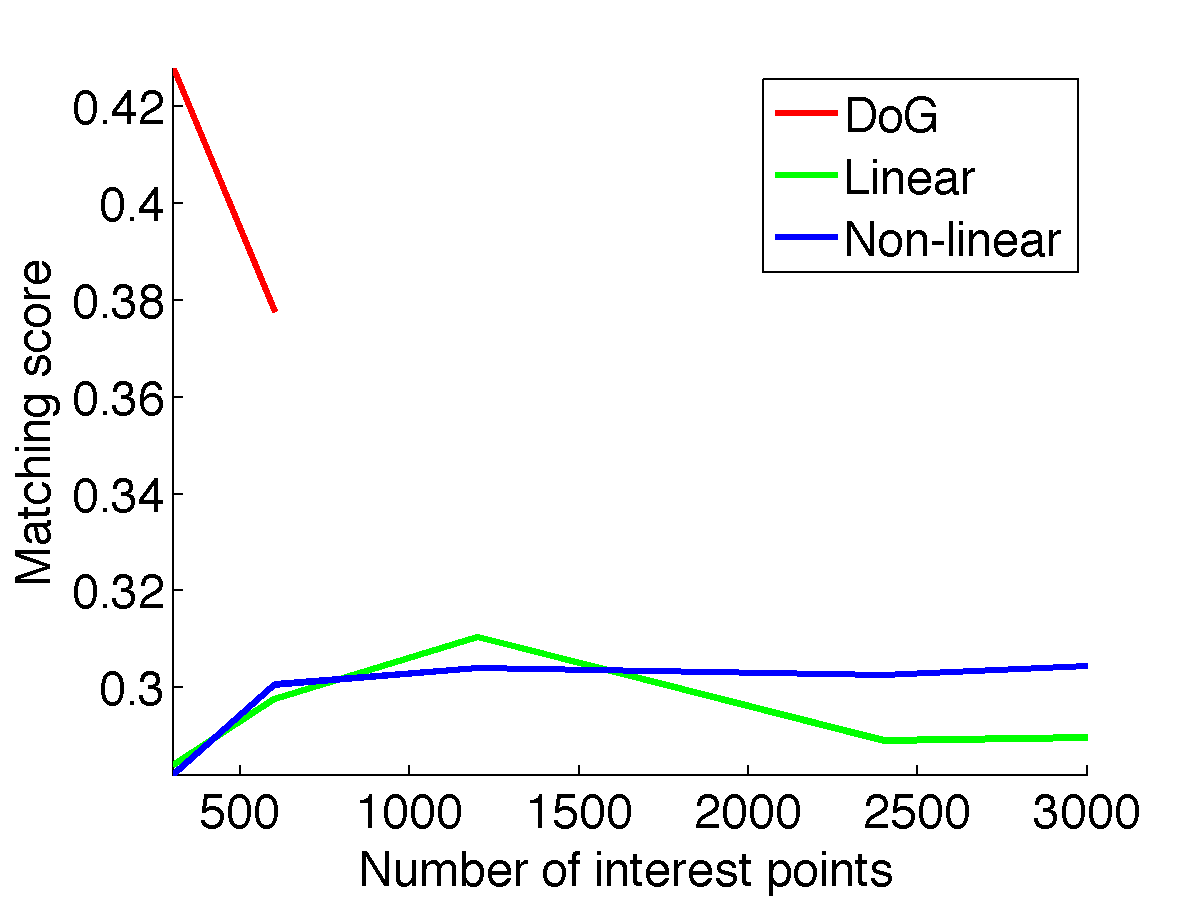} \\
 Trees (blur) & UBC (jpeg)
 \end{tabular}
 \caption{\label{fig:matching} Matching score (the higher, the better) of DoG and our methods (Linear, Non-linear) on the benchmark from~\cite{mikolajczyk2005comparison}.}
 \end{figure}

Finally, we combine our detector with the SIFT descriptor and measure how well the detected points can be matched. For that we use the same matching score as in~\cite{mikolajczyk2005comparison}, i.e., the ratio of correct matches to all matches. Our detectors (Linear, Non-linear)+SIFT are slightly better than DoG+SIFT in most cases, as shown in Fig.~\ref{fig:matching}. Our methods performed worse than DoG only for the UBC dataset, measuring the robustness to the JPEG compression, which was not included in our training.

\def \wf {0.05}
\begin{figure*}
 \centering
 \begin{tabular}{ccc}
  \includegraphics[width=\wf\linewidth]{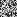} &
 \includegraphics[width=\wf\linewidth]{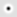} &
 \includegraphics[width=\wf\linewidth]{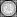} \\
random & DoG & ours linear
 \end{tabular}
 \caption{\label{fig:filters} Filters of linear models. The DoG filter parameters default to the standard implementation~\cite{opencv_library}.}
 \end{figure*}
 
\def \wlf {0.2}
\begin{figure*}
 \centering
 \begin{tabular}{ccccc}
  \includegraphics[width=\wlf\linewidth]{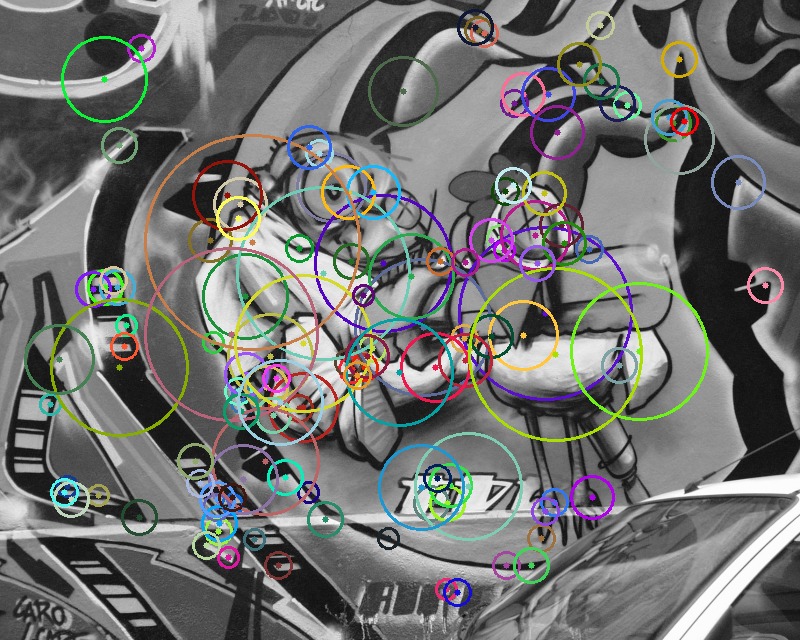} &
  \includegraphics[width=\wlf\linewidth]{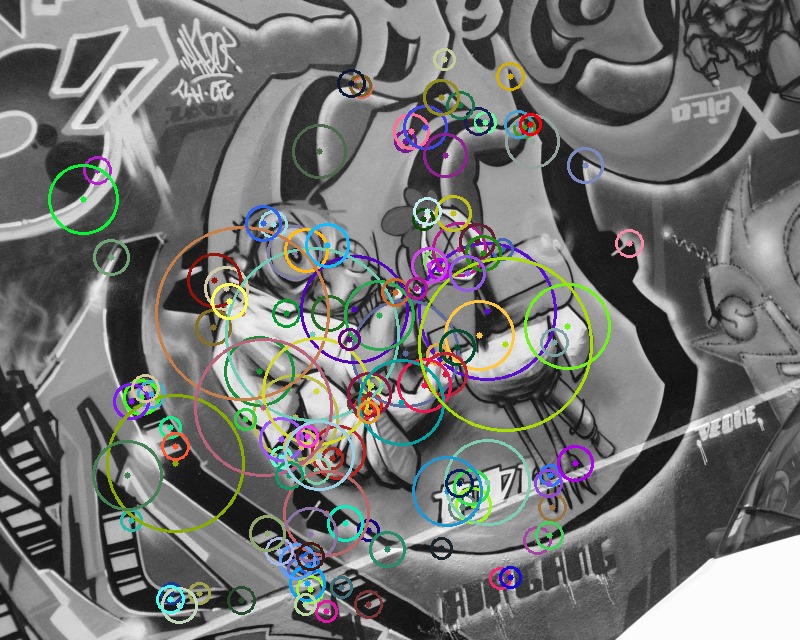} & &
   \includegraphics[width=\wlf\linewidth]{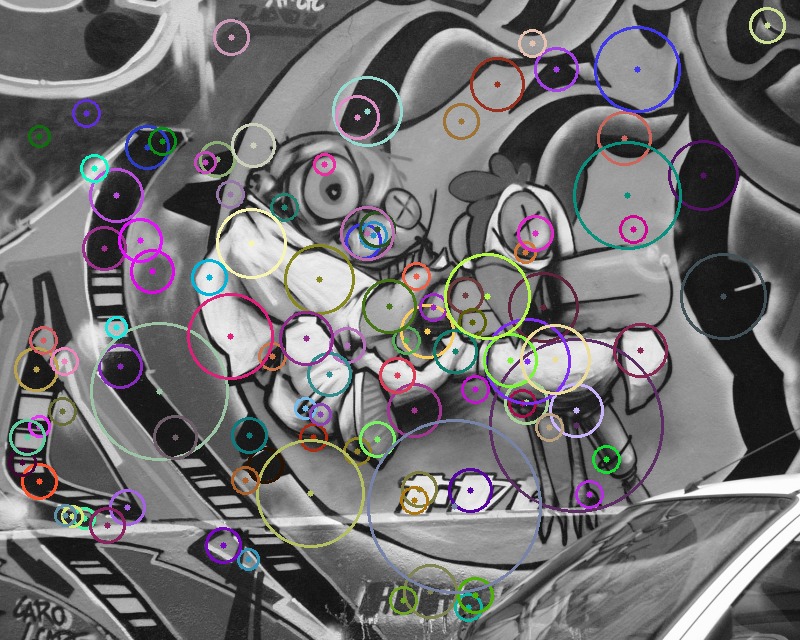} &
  \includegraphics[width=\wlf\linewidth]{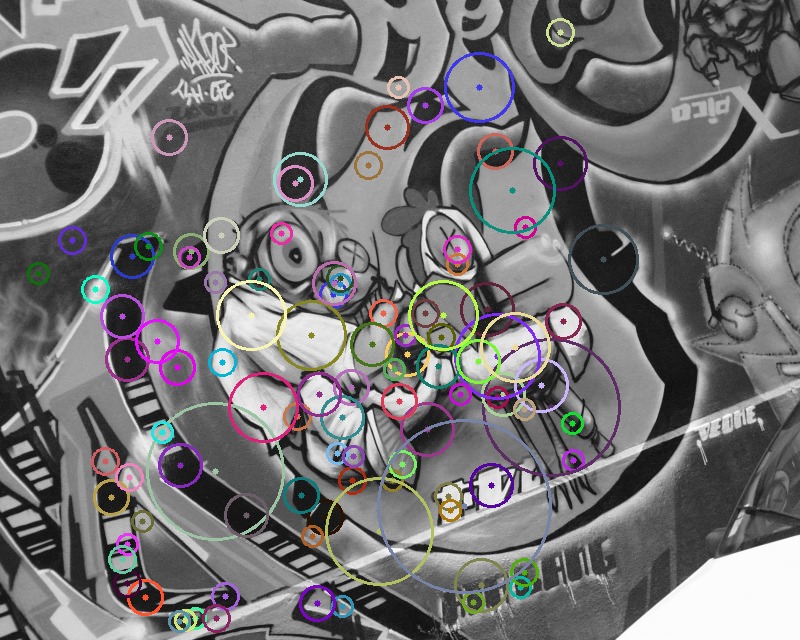} \\
  \includegraphics[width=\wlf\linewidth]{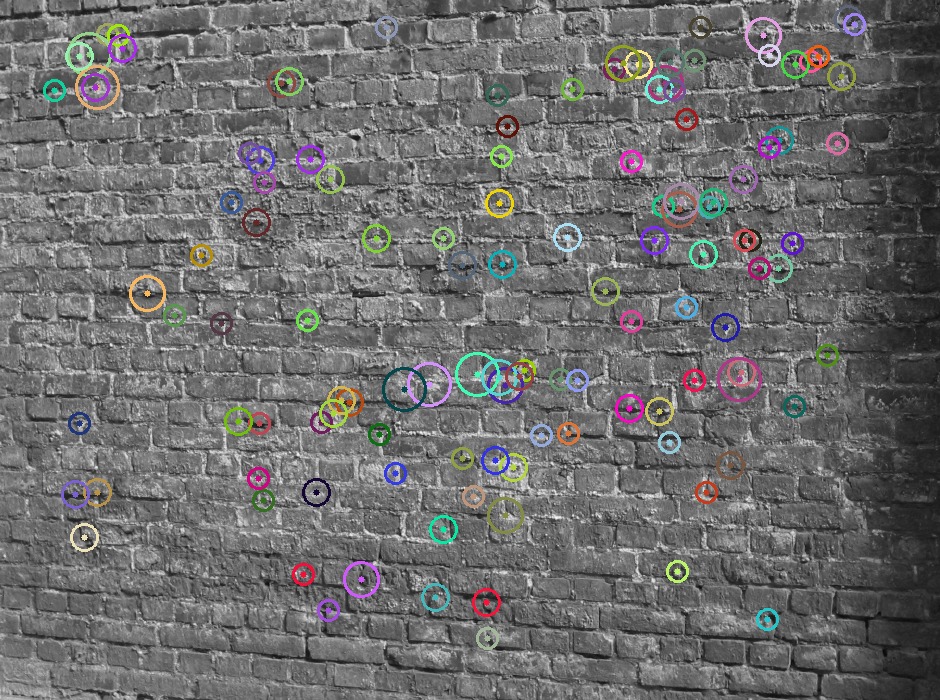} &
  \includegraphics[width=\wlf\linewidth]{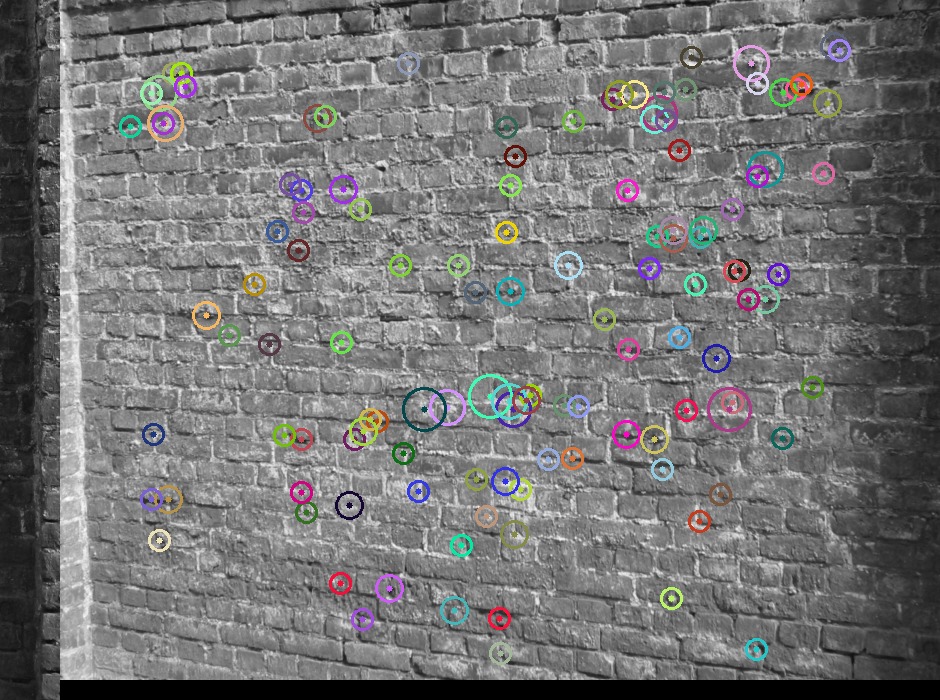} & &
   \includegraphics[width=\wlf\linewidth]{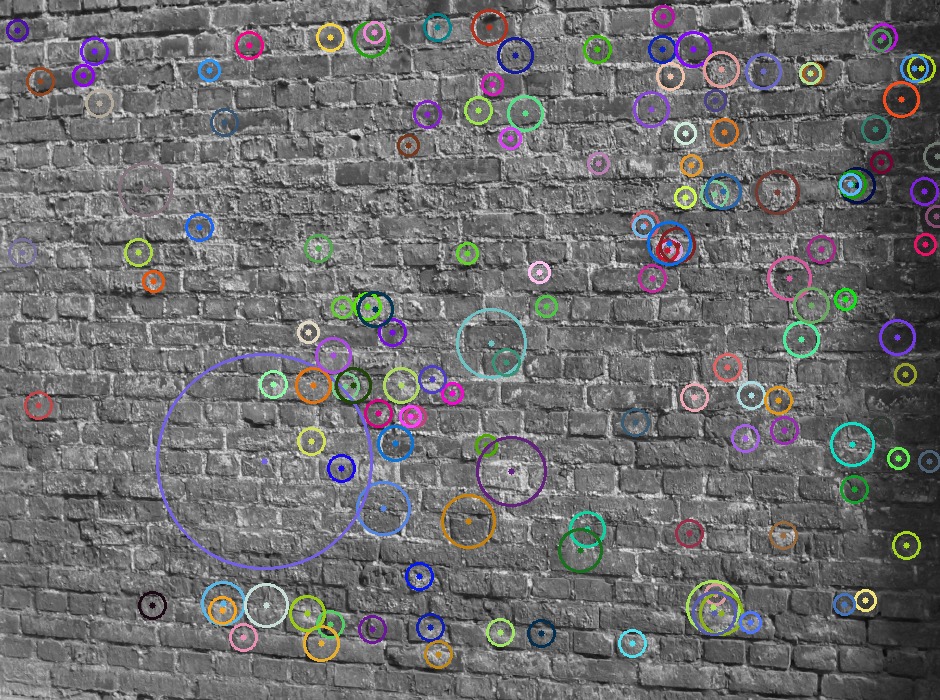} &
  \includegraphics[width=\wlf\linewidth]{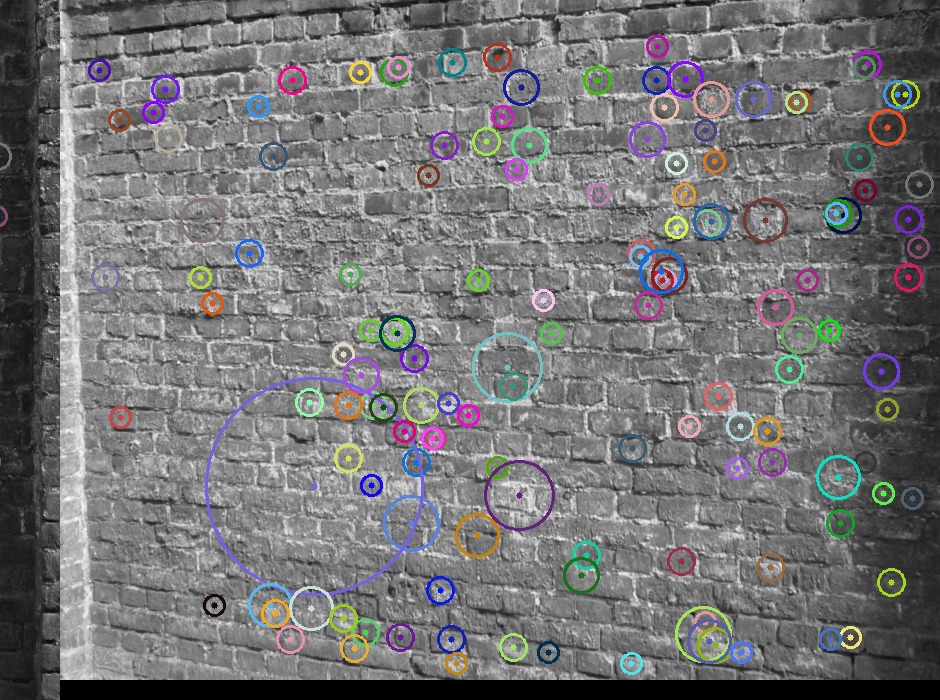} \\
  \includegraphics[width=\wlf\linewidth]{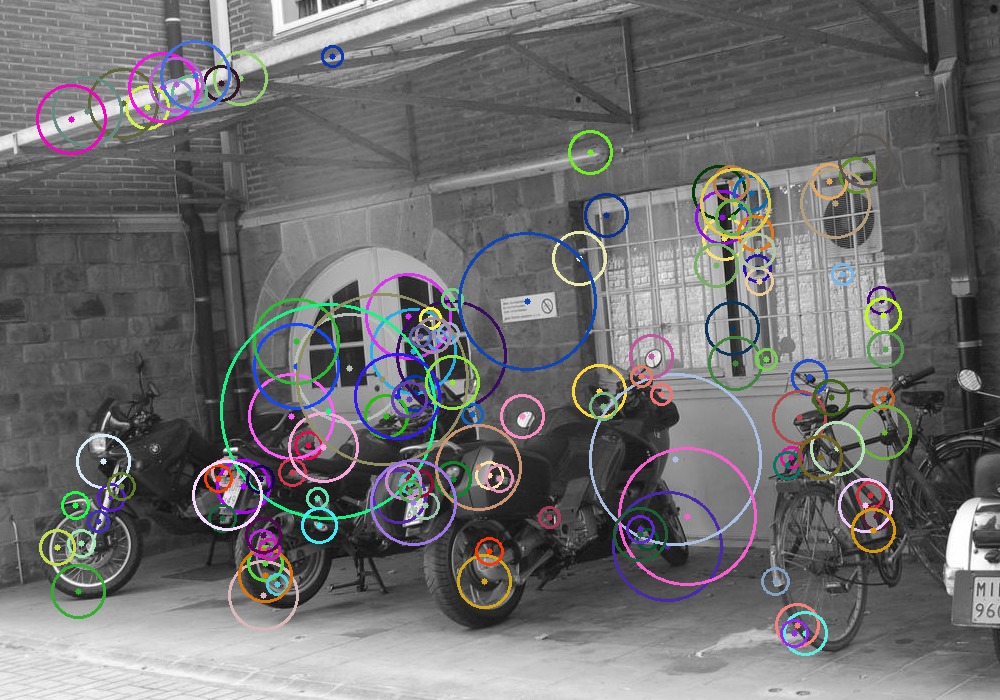} &
  \includegraphics[width=\wlf\linewidth]{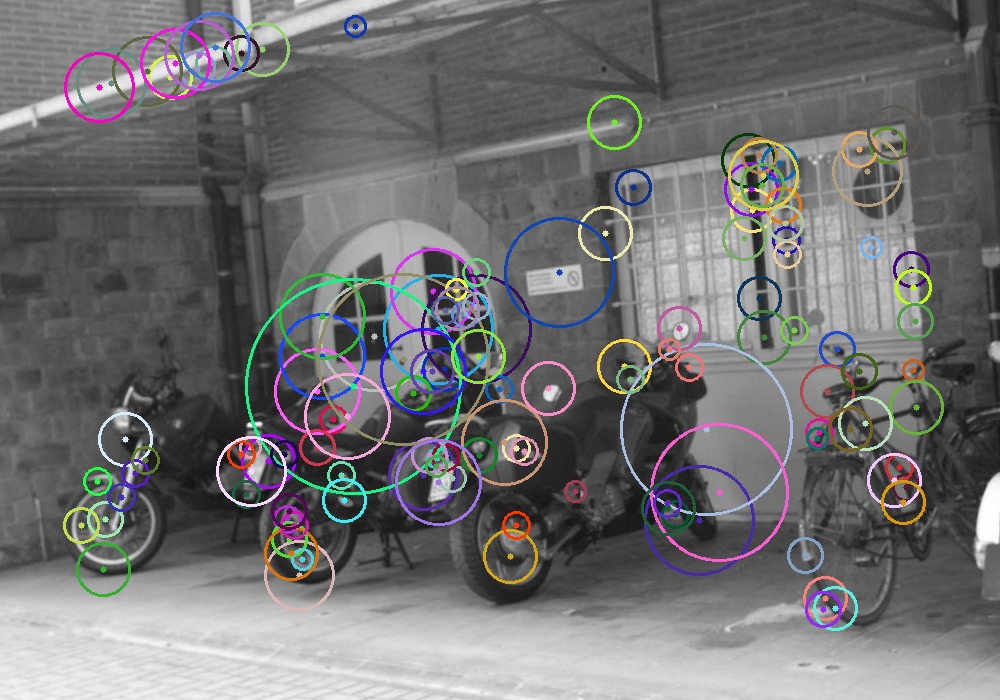} & &
   \includegraphics[width=\wlf\linewidth]{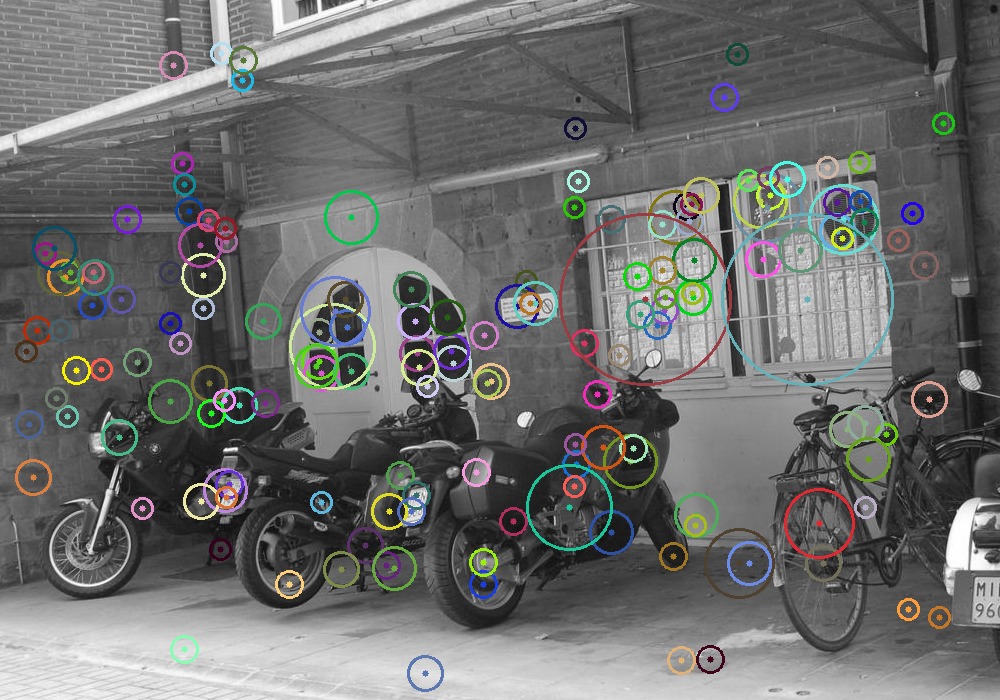} &
  \includegraphics[width=\wlf\linewidth]{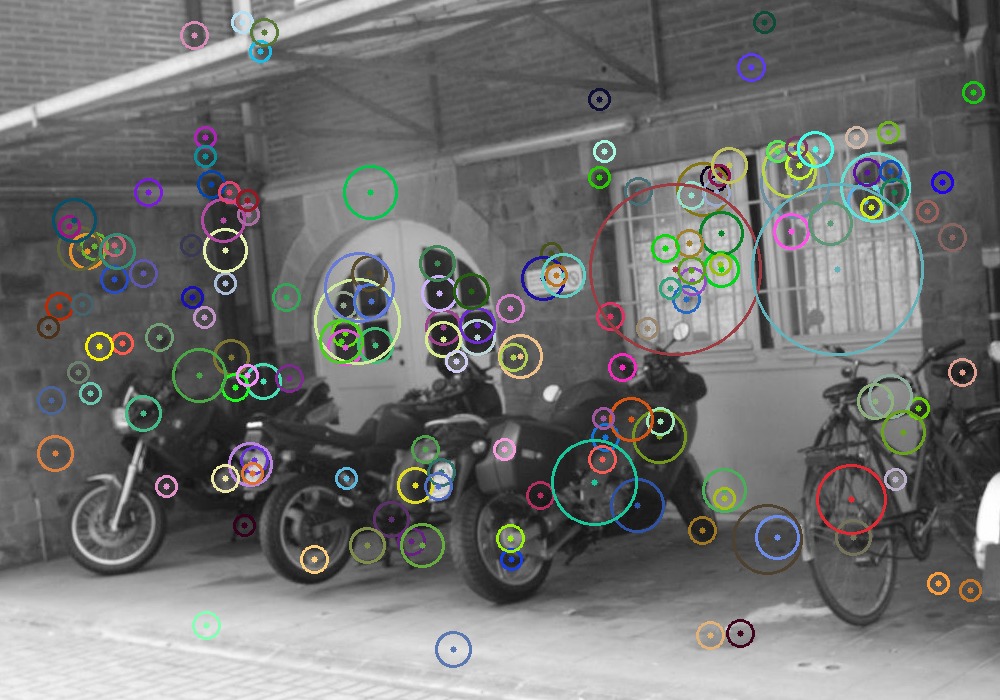} \\
  \includegraphics[width=\wlf\linewidth]{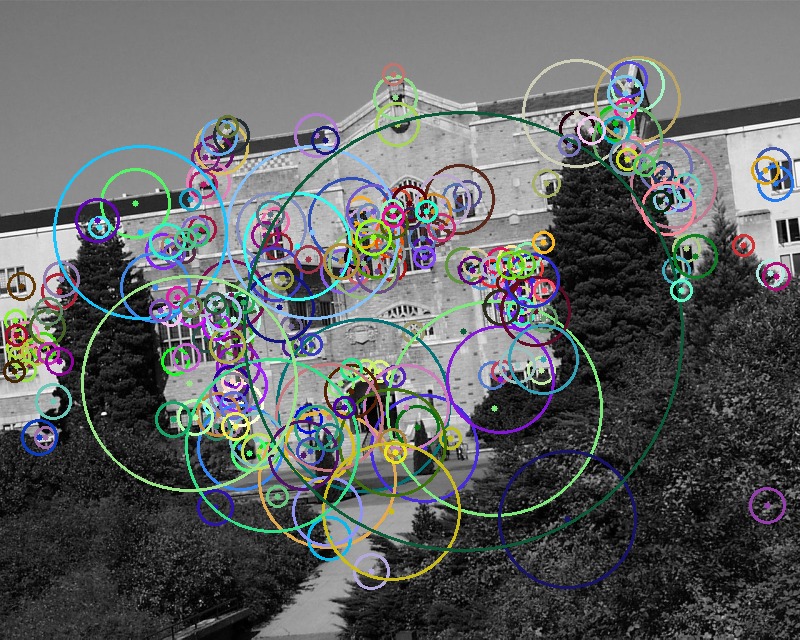} &
  \includegraphics[width=\wlf\linewidth]{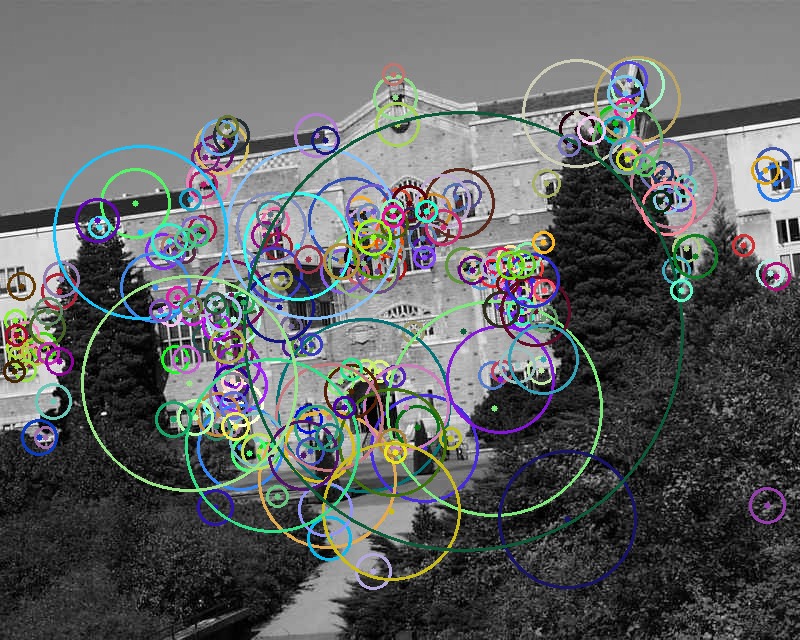} & &
   \includegraphics[width=\wlf\linewidth]{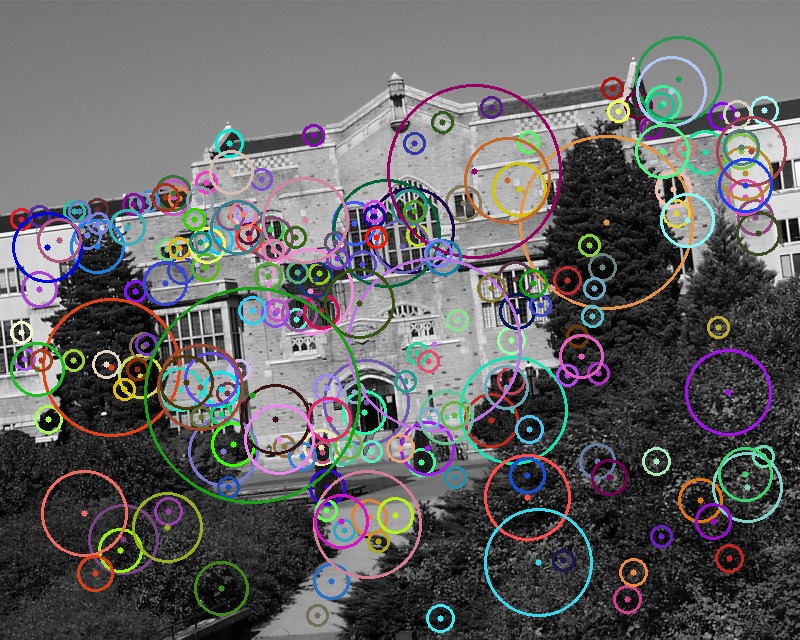} &
  \includegraphics[width=\wlf\linewidth]{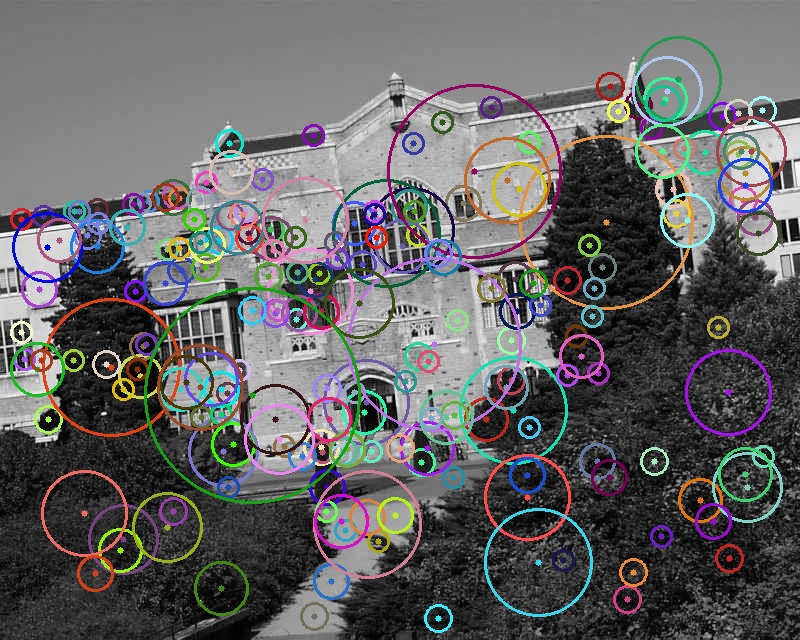} \\
    DoG left & DoG right & & ours left & ours right
   \end{tabular}
 \caption{\label{fig:detections} Correct (repeatable) detections. Rows correspond to datasets in the following order: graf1-2, wall1-2, bikes1-2, ubc1-2.}
 \end{figure*}
 

\begin{table}[]
\centering
\caption{Repeatability of a random filter, DoG, and our linear (Linear) and non-linear (Non-lin) methods. Some entries are omitted because of having not enough points after non-maximum suppression. The left-most column is the transformation class, we used abbreviations: VP for viewpoint, Z+R for zoom+rotation, L for illumination.}
\label{tab:3d}
\footnotesize{
\begin{tabular}{@{}llllllll@{}}
\toprule
      &        &         & \multicolumn{5}{c}{Number of interest points}                                 \\ \midrule
T     & Data   & Method  & 300           & 600           & 1200          & 2400          & 3000          \\ \midrule
VP  & graf   & Random  & 0.06          & 0.08          & 0.12          & 0.17          & 0.19          \\
      &        & DoG     & \textbf{0.21} & \textbf{0.2}  & 0.18          & -             & -             \\
      &        & Linear  & 0.17          & 0.18          & 0.19          & 0.21          & 0.22          \\
      &        & Non-lin & 0.17          & 0.19          & \textbf{0.21} & \textbf{0.24} & \textbf{0.25} \\ \cmidrule(l){3-8} 
      & wall   & Random  & 0.18          & 0.22          & 0.27          & 0.33          & 0.36          \\
      &        & DoG     & 0.27          & 0.28          & 0.28          & -             & -             \\
      &        & Linear  & \textbf{0.33} & \textbf{0.36} & \textbf{0.39} & 0.43          & 0.44          \\
      &        & Non-lin & 0.3           & 0.35          & \textbf{0.39} & \textbf{0.44} & \textbf{0.46} \\ \cmidrule(l){3-8} 
Z+R   & bark   & Random  & 0.02          & 0.03          & 0.05          & 0.08          & 0.1           \\
      &        & DoG     & 0.13          & 0.13          & -             & -             & -             \\
      &        & Linear  & \textbf{0.14} & \textbf{0.15} & \textbf{0.15} & 0.15          & -             \\
      &        & Non-lin & 0.12          & 0.13          & 0.14          & \textbf{0.16} & \textbf{0.16} \\ \cmidrule(l){3-8} 
      & boat   & Random  & 0.03          & 0.05          & 0.08          & 0.11          & 0.12          \\
      &        & DoG     & 0.26          & 0.25          & 0.2           & -             & -             \\
      &        & Linear  & \textbf{0.27} & \textbf{0.27} & 0.27          & 0.26          & 0.25          \\
      &        & Non-lin & 0.21          & 0.24          & \textbf{0.28} & \textbf{0.28} & \textbf{0.29} \\ \cmidrule(l){3-8} 
L & leuven & Random  & 0.51          & 0.57          & 0.63          & 0.69          & 0.71          \\
      &        & DoG     & 0.51          & 0.51          & 0.5           & -             & -             \\
      &        & Linear  & 0.69          & 0.69          & 0.73          & 0.73          & 0.72          \\
      &        & Non-lin & \textbf{0.7}  & \textbf{0.72} & \textbf{0.75} & \textbf{0.76} & \textbf{0.77} \\ \cmidrule(l){3-8} 
Blur  & bikes  & Random  & 0.36          & 0.42          & 0.48          & 0.53          & 0.54          \\
      &        & DoG     & 0.41          & 0.41          & 0.39          & -             & -             \\
      &        & Linear  & \textbf{0.53} & \textbf{0.53} & 0.49          & \textbf{0.55} & \textbf{0.57} \\
      &        & Non-lin & 0.52          & 0.51          & \textbf{0.51} & 0.49          & 0.49          \\ \cmidrule(l){3-8} 
      & trees  & Random  & 0.21          & 0.26          & 0.32          & 0.4           & 0.43          \\
      &        & DoG     & 0.29          & 0.3           & 0.31          & -             & -             \\
      &        & Linear  & 0.34          & 0.37          & 0.42          & 0.45          & \textbf{0.5}  \\
      &        & Non-lin & \textbf{0.36} & \textbf{0.39} & \textbf{0.44} & \textbf{0.49} & \textbf{0.5}  \\ \cmidrule(l){3-8} 
JPEG  & ubc    & Random  & 0.42          & 0.47          & 0.53          & 0.59          & 0.61          \\
      &        & DoG     & \textbf{0.68} & 0.6           & -             & -             & -             \\
      &        & Linear  & 0.55          & \textbf{0.62} & \textbf{0.66} & 0.67          & 0.68          \\
      &        & Non-lin & 0.58          & \textbf{0.62} & 0.64          & \textbf{0.69} & \textbf{0.7}  \\ \bottomrule
\end{tabular}
}
\end{table}

\subsection{Fully-unsupervised RGB detector}
The goal of this experiment is to show that ground-truth correspondences from an additional data source (like 3D points from a laser scanner) are not necessary to train a detector with our method. Instead, we can sample random transformations to obtain correspondences.

\noindent\textbf{Training.} In this experiment, we only used images from the DTU dataset with different illuminations. To generate the correspondence, a patch was randomly selected from an image and randomly transformed. We considered affine warps, preserving area, together with illumination changes, uniformly sampled from those provided by the dataset. The affine warps were parametrized as $rot(\alpha) * diag(s, \frac{1}{s} ) * rot(-\alpha))$ by a rotation $\alpha$ (uniformly sampled from $[0, 2 \pi]$) and a scaling factor $s$. We considered two settings: small warps ($s$ sampled uniformly from $[1, 1.1]$) and large warps ($s$ sample uniformly from $[1, 2]$).

\noindent\textbf{Testing.} We used the Oxford VGG dataset~\cite{mikolajczyk2005comparison} (same as in the previous experiment).

\noindent\textbf{NN architectures.} We considered linear models.

\noindent\textbf{Results.} As shown in Table~\ref{tab:warp}, our methods outperform DoG in more than half of the cases.


\begin{figure}[!ht]
 \centering
 \includegraphics[width=\linewidth]{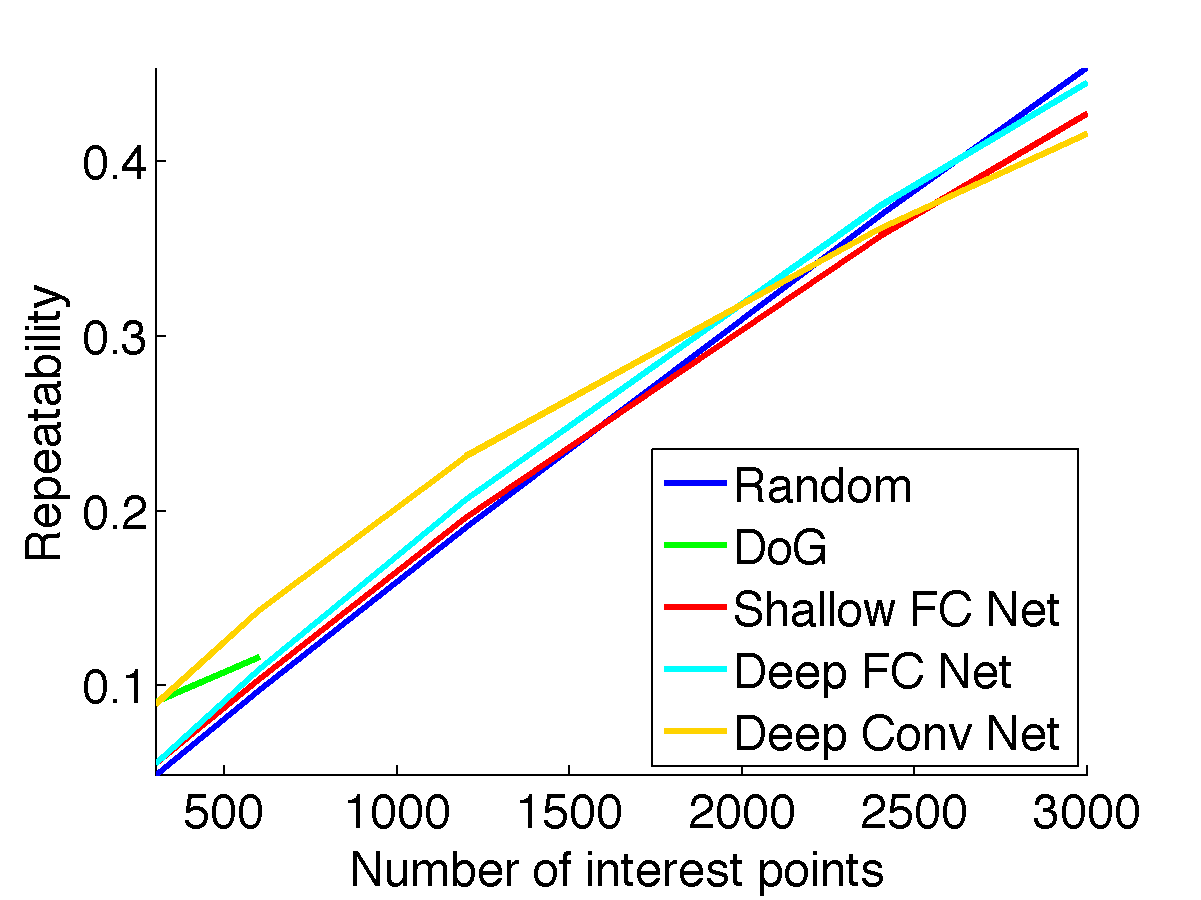}
 \caption{\label{fig:nyu_rep} Our deep convolutional model (Deep Conv Net) produces overall better repeatability than baselines.}
 \end{figure}

\def \wlf {0.18}
\begin{figure*}
 \centering
 \begin{tabular}{ccccc}
 \includegraphics[width=\wlf\linewidth]{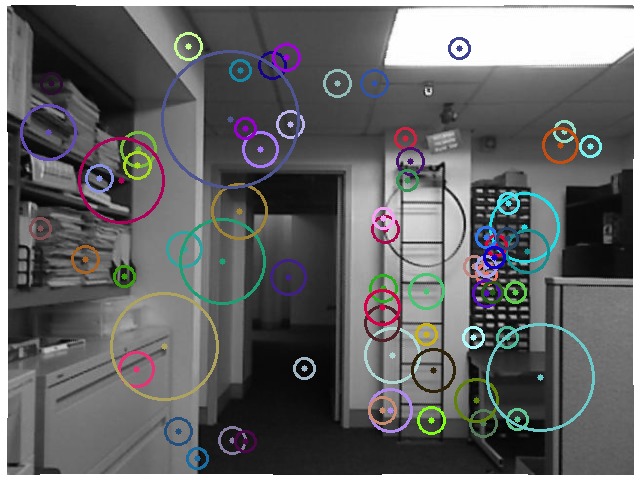} &
 \includegraphics[width=\wlf\linewidth]{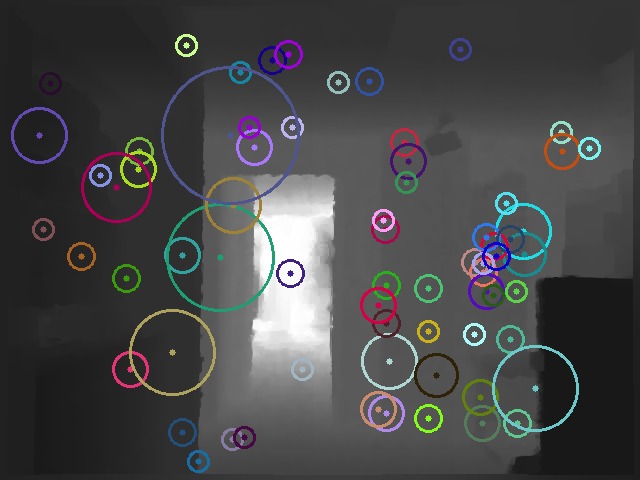} & &
 \includegraphics[width=\wlf\linewidth]{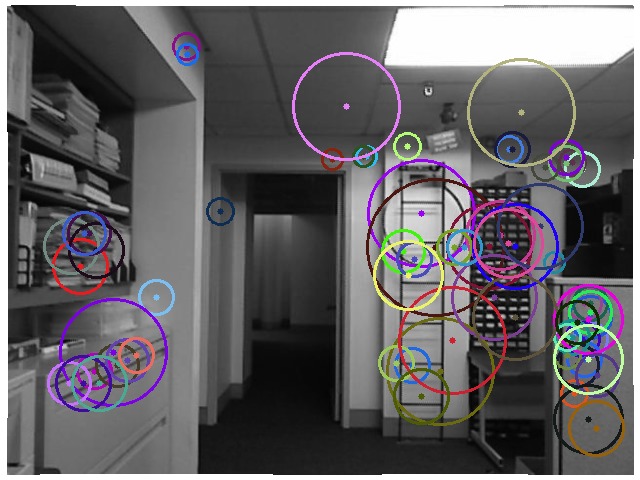} &
 \includegraphics[width=\wlf\linewidth]{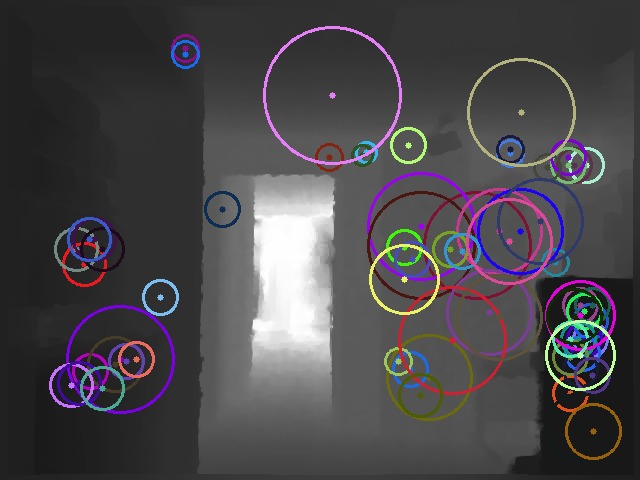} \\
 \includegraphics[width=\wlf\linewidth]{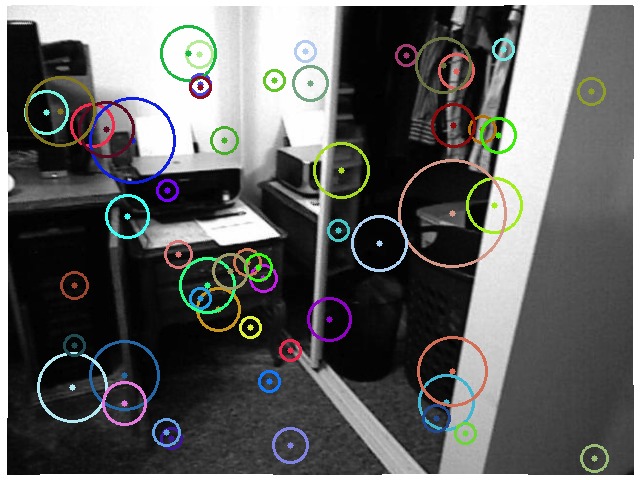} &
 \includegraphics[width=\wlf\linewidth]{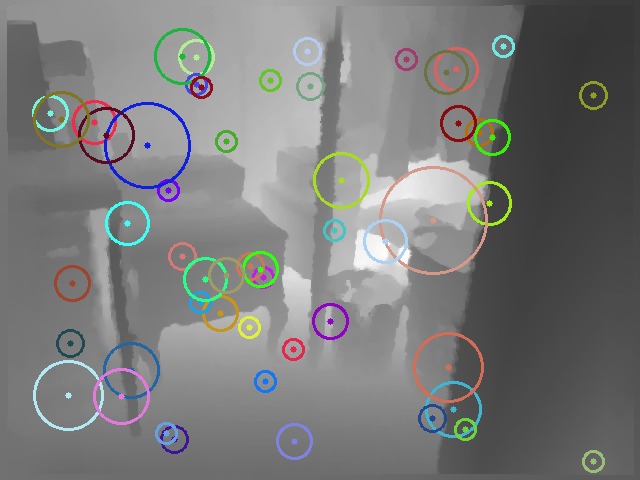} & &
  \includegraphics[width=\wlf\linewidth]{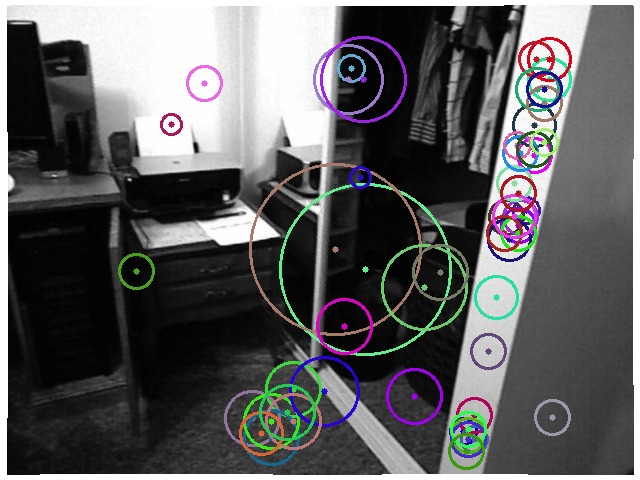} &
 \includegraphics[width=\wlf\linewidth]{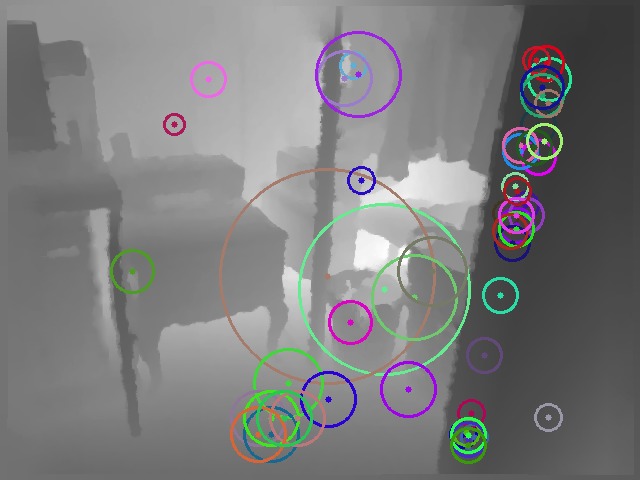} \\
\includegraphics[width=\wlf\linewidth]{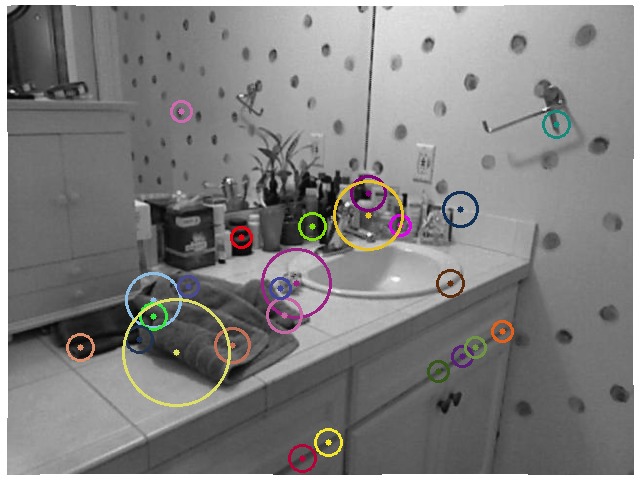} &
 \includegraphics[width=\wlf\linewidth]{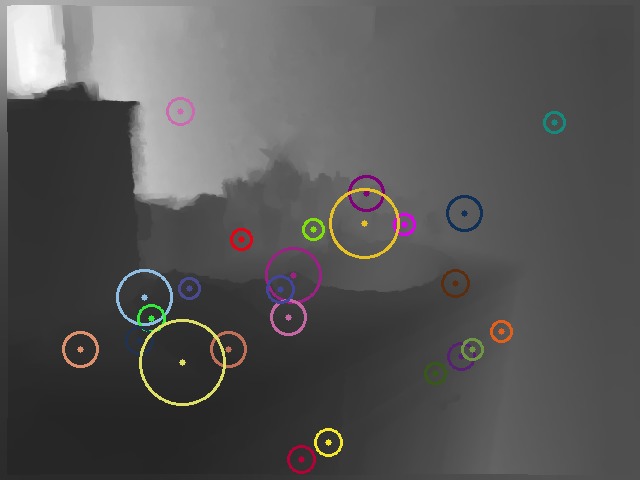} & &
  \includegraphics[width=\wlf\linewidth]{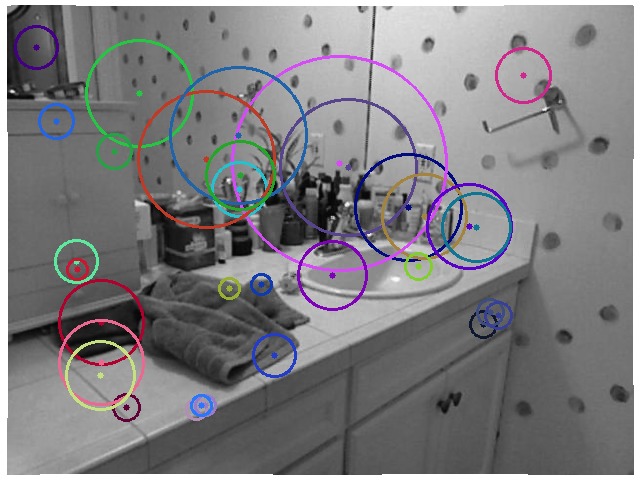} &
 \includegraphics[width=\wlf\linewidth]{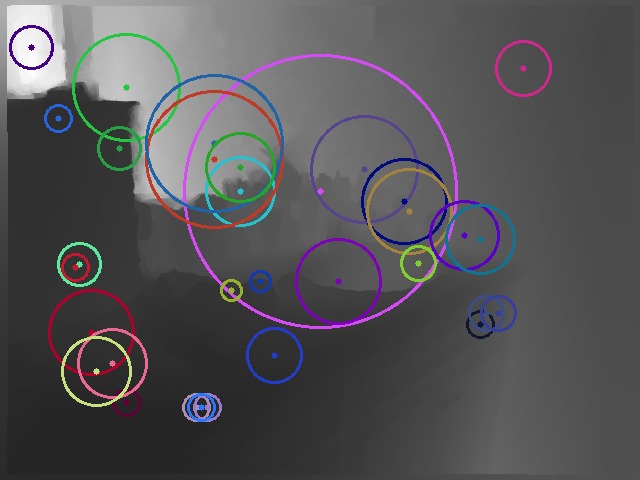} \\
 \includegraphics[width=\wlf\linewidth]{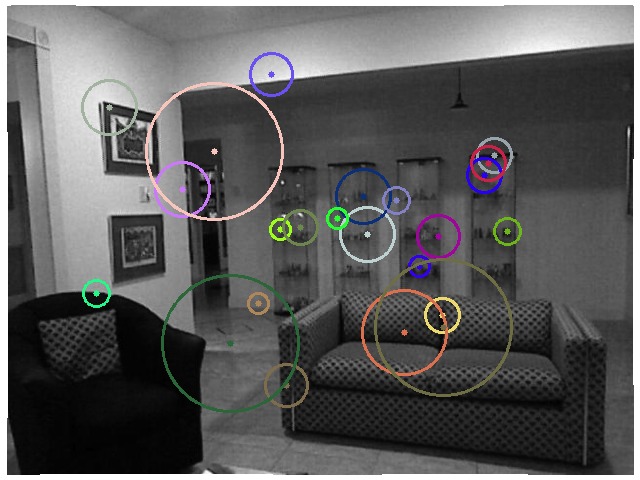} &
 \includegraphics[width=\wlf\linewidth]{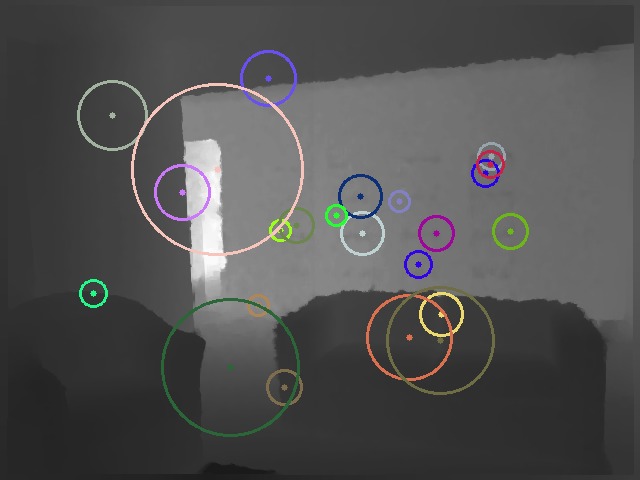} & &
  \includegraphics[width=\wlf\linewidth]{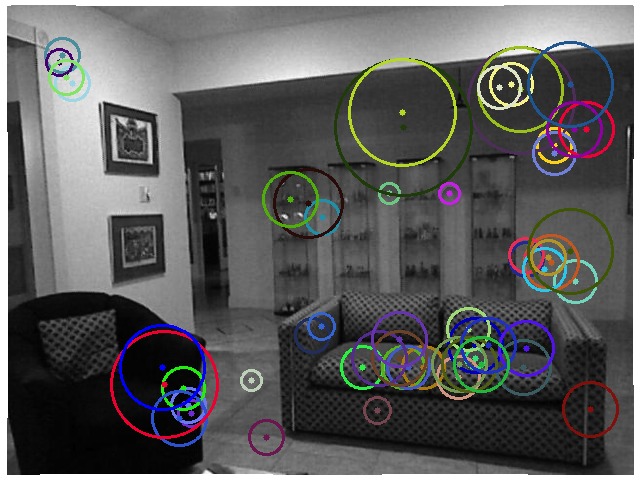} &
 \includegraphics[width=\wlf\linewidth]{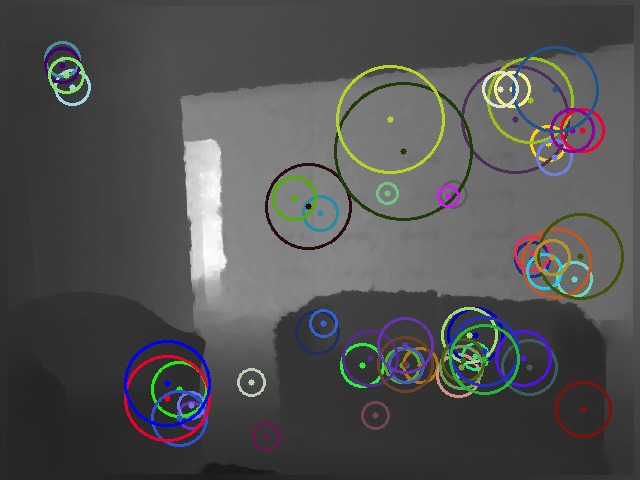} \\
 DoG image & DoG depth & & ours image & ours depth
 \end{tabular}
 \caption{\label{fig:nyu_response} Correct (repeatable) detections. Rows correspond to frames 17, 529, 717, 1257 from NYUv2.}
 \end{figure*}

\begin{table}[!h]
\centering
\caption{Repeatability of DoG, our methods learned with large (WarpL) and small (WarpS) warps.}
\label{tab:warp}
\footnotesize{
\begin{tabular}{@{}llllllll@{}}
\toprule
      &        &        & \multicolumn{5}{c}{Number of interest points}                                 \\ \midrule
T     & Data   & Method & 300           & 600           & 1200          & 2400          & 3000          \\ \midrule
VP  & graf   & DoG    & \textbf{0.21} & \textbf{0.2}  & \textbf{0.18} & -             & -             \\
      &        & WarpL  & 0.15          & 0.15          & 0.17          & 0.18          & 0.19          \\
      &        & WarpS  & 0.14          & 0.17          & \textbf{0.18} & \textbf{0.19} & \textbf{0.2}  \\ \cmidrule(l){3-8} 
      & wall   & DoG    & 0.27          & 0.28          & 0.28          & -             & -             \\
      &        & WarpL  & \textbf{0.35} & \textbf{0.37} & \textbf{0.39} & \textbf{0.42} & \textbf{0.42} \\
      &        & WarpS  & 0.27          & 0.32          & 0.36          & 0.41          & \textbf{0.42} \\ \cmidrule(l){3-8} 
Z+R   & bark   & DoG    & \textbf{0.13} & \textbf{0.13} & -             & -             & -             \\
      &        & WarpL  & 0.09          & 0.09          & 0.09          & -             & -             \\
      &        & WarpS  & 0.11          & 0.12          & \textbf{0.13} & \textbf{0.14} & -             \\ \cmidrule(l){3-8} 
      & boat   & DoG    & \textbf{0.26} & \textbf{0.25} & 0.2           & -             & -             \\
      &        & WarpL  & 0.16          & 0.18          & 0.18          & 0.19          & 0.19          \\
      &        & WarpS  & 0.2           & 0.21          & \textbf{0.22} & \textbf{0.22} & \textbf{0.23} \\ \cmidrule(l){3-8} 
L & leuven & DoG    & 0.51          & 0.51          & 0.5           & -             & -             \\
      &        & WarpL  & 0.66          & 0.64          & 0.65          & 0.67          & 0.67          \\
      &        & WarpS  & \textbf{0.69} & \textbf{0.67} & \textbf{0.68} & \textbf{0.71} & \textbf{0.71} \\ \cmidrule(l){3-8} 
Blur  & bikes  & DoG    & 0.41          & 0.41          & 0.39          & -             & -             \\
      &        & WarpL  & 0.49          & 0.46          & 0.42          & 0.52          & -             \\
      &        & WarpS  & \textbf{0.55} & \textbf{0.54} & \textbf{0.52} & \textbf{0.57} & \textbf{0.6}  \\ \cmidrule(l){3-8} 
      & trees  & DoG    & 0.29          & 0.3           & 0.31          & -             & -             \\
      &        & WarpL  & 0.31          & 0.35          & 0.38          & 0.43          & 0.47          \\
      &        & WarpS  & \textbf{0.33} & \textbf{0.37} & \textbf{0.41} & \textbf{0.44} & \textbf{0.49} \\ \cmidrule(l){3-8} 
JPEG  & ubc    & DoG    & \textbf{0.68} & \textbf{0.6}  & -             & -             & -             \\
      &        & WarpL  & 0.54          & 0.59          & 0.61          & 0.61          & 0.62          \\
      &        & WarpS  & 0.54          & \textbf{0.6}  & \textbf{0.65} & \textbf{0.67} & \textbf{0.67} \\ \bottomrule
\vspace{-4em}
\end{tabular}
}
\end{table}

\subsection{Cross-modal RGB/depth detector}
In this experiment, we show how to use our method for learning a cross-modal detector --- a hard problem where we do not have an understanding on how to design a good solution by hand. We learn a detector between RGB and depth images by training on the NYUv2 dataset~\cite{Silberman:ECCV12}. Such a detector has an application in augmenting an un-colored 3D point cloud with colors from a newly obtained image.

\noindent\textbf{Training.} We use $40$ random frames from NYUv2, which contains view-aligned Kinect RGBD frames (an RGB pixel corresponds to a depth pixel at the same location). 

\noindent\textbf{Testing.} We use $40$ random frames from NYUv2 (unrelated to the training set).

\noindent\textbf{NN architectures.}
We evaluated the following architectures for the response function $H$:
\begin{itemize}
	\item Deep convolutional network (Deep Conv Net): $(c(7, 1, 32, 3), b, e, (c(7, 32, 32, 3), b, e)^8, c(17, 32, 1, 0))$, 
	\item Shallow fully-connected network (Shallow FC Net): $(c(17, 1, 32, 0), e, f(32, 32), e, f(32, 1))$,
	\item Deep fully-connected network (Deep FC Net): $(c(17, 1, 32, 0), e, (f(32, 32), e)^8, f(32, 1))$.
\end{itemize}

\begin{figure}[!ht]
 \centering
  \includegraphics[width=0.67\linewidth]{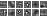}
  \caption{\label{fig:nyu_filters} Some $7$x$7$ filters from the first layer of our deep convolutional model (Deep Conv Net), it is possible to see edge-like filters, blob filters and high-frequency filters.}
 \end{figure}

\noindent\textbf{Results.} The repeatability and filters from the best model (Deep Conv Net) are shown in Fig.~\ref{fig:nyu_rep} and Fig.~\ref{fig:nyu_filters}. Our best model outperformes others by a large relative value. As shown in the repeatability plot, DoG produces a relatively small number of interest points. That is because we extract the same number of points from both sensors --- for the fair comparison as explained at the beginning of the section --- and DoG produces very few of them (after non-maximum suppression) in the depth channel, which is very smooth and lacks texture. On the opposite, our methods produce more points as they learn to "spread" image patches during training, making the response distribution more peaky. We compare the detections of our best model to DoG in Fig.~\ref{fig:nyu_response}.
 
\section{Conclusion}
\label{sec:conclusion}
In this work, we have proposed an unsupervised approach to learning an interest point detector. The key idea of the method is to produce a repeatable ranking of points of the object and use top/bottom quantiles of the ranking as interest points. We have demonstrated how to learn such a detector for images. We show superior or comparable performance of our method with respect to DoG in two different settings: learning standard RGB detector from scratch and learning a detector, repeatable between different modalities (RGB and depth from Kinect). Future work includes learning the descriptor jointly with our detector. Also, one could investigate applying our method to detection beyond images (\eg, to interest frame detection in videos).

\textbf{Acknowledgements:}
This work is partially funded by the Swiss NSF project 163910, the Max Planck CLS Fellowship and the Swiss CTI project 17136.1 PFES-ES. 

\newpage
{\small
\bibliographystyle{ieee}
\bibliography{cvpr_2017}
}

\end{document}